\documentclass{article}

\PassOptionsToPackage{numbers,sort&compress}{natbib}
\usepackage[preprint]{neurips_2026}

\usepackage[utf8]{inputenc}
\usepackage[T1]{fontenc}
\usepackage[hypertexnames=false]{hyperref}
\usepackage{url}
\usepackage{graphicx}
\usepackage{booktabs}
\usepackage{multirow}
\usepackage{threeparttable}
\usepackage[table]{xcolor}
\usepackage{amssymb}
\usepackage{algorithm}
\usepackage{algpseudocode}
\usepackage{amsthm}
\usepackage[shortlabels]{enumitem}
\usepackage[section]{placeins}

\usepackage{amsmath,amsfonts,bm}

\def\eqref#1{equation~\ref{#1}}
\def\1{\bm{1}}

\DeclareMathAlphabet{\mathsfit}{\encodingdefault}{\sfdefault}{m}{sl}
\SetMathAlphabet{\mathsfit}{bold}{\encodingdefault}{\sfdefault}{bx}{n}

\newcommand{\E}{\mathbb{E}}

\DeclareMathOperator*{\argmin}{arg\,min}

\makeatletter
\renewcommand{\@notice}{}
\makeatother

\newcommand{\ours}{RAPS-DA}

\title{Regime-Aware Peer Specialization for Robust RAG under Heterogeneous Knowledge Conflicts}

\author{%
  Bo Wang \quad Heyan Huang \quad Yaolin Li \quad Yanghao Zhou \\
  \textbf{Jiahao Teng} \quad \textbf{Ziyi Yang} \quad \textbf{Ge Shi} \quad \textbf{Chong Feng} \\
  School of Computer Science,  Beijing Institute of Technology
}

\begin{document}

\maketitle

\begin{abstract}
Retrieval-augmented generation (RAG) improves language models by grounding generation in external context.
However, it can be fragile when the retrieved context conflicts with the model's parametric knowledge.
Such conflicts span a reliability spectrum, ranging from reliable and partially reliable evidence to adversarial context.
Existing remedies often handle such heterogeneous conflicts with regime-agnostic supervision, which can conflate incompatible learning signals across reliability regimes.
To disentangle these signals, we propose \textbf{\ours{}}, a regime-aware peer specialization framework that addresses conflict at two complementary granularities.
At the \emph{sample level}, conflicts are divided into three regimes, including \textbf{Grounding}, \textbf{Arbitration}, and \textbf{Resistance}, with one same-scale peer specialist trained per regime from a shared base model.
Each sample is then hard-routed to its regime-matched peer for on-policy reverse-KL supervision.
At the \emph{token level}, a dual-layer selector uses inter-teacher disagreement, student--teacher divergence, and student entropy to filter uninformative or unstable tokens, upweight confidently misaligned ones, and gradually focus supervision on high-conflict tokens as the student matures.
Gains stem from specialization at a fixed model scale, not from a stronger teacher, and the peer specialists exist only during training, so the deployed student requires no regime labels or peer access.
Experiments on five conflict scenarios and two out-of-distribution benchmarks show \ours{} surpasses all prompting, decoding, fine-tuning, RL, and single-teacher baselines.
% On Qwen-7B, it achieves the largest gains in the Resistance regime, exceeding the best RL baseline by 11.2 points, and a single 7B student closes 69.6\% of the gap to an oracle regime-matched specialist.
\end{abstract}

\section{Introduction}\label{sec:intro}

Retrieval-augmented generation (RAG) grounds large language models in external evidence, yet its success rests on the assumption that retrieved context is reliable~\citep{lewis2020retrieval,guu2020retrieval}.
When this assumption breaks, RAG systems face a trust--resistance dilemma: they may over-trust incorrect passages~\citep{su2024conflictbank,Xie2024KnowledgeConflict}, while resistance-oriented training can cause them to reject genuinely informative evidence~\citep{bi2025parametersvscontextfinegrained,wang2025ramdocs}.
Generally, context reliability is a spectrum in which reliable evidence should be integrated, adversarial evidence should be rejected, and mixed-quality evidence requires selective trust.
% This raises a central question.
This heterogeneity creates a training challenge: \emph{How can a single model be trained when different reliability regimes demand conflicting behaviors?}

A growing body of work has sought to address this challenge.
Prompting and decoding methods~\citep{ding2024retrieve,shi2024cad,chuang2024dola} improve inference-time behavior but generalize poorly.
Supervised and RL methods~\citep{asai2024self,fang2024raat,lin2025knowledgeable} apply a single training signal across mixed-reliability data, so conflicting regimes impose contradictory gradient updates.
These methods either lack a trainable adaptation mechanism or provide supervision that is too coarse to resolve token-level follow-or-reject decisions.
By contrast, on-policy distillation (OPD)~\citep{agarwal2024gkd,gu2024minillm} offers a more suitable foundation: it provides dense token-level supervision on the student's own rollouts, directly shaping pivotal tokens where the model decides whether to follow or reject a retrieved claim~\citep{fu2026revisiting,tip2026}. 
Yet naive OPD still uses one teacher for all regimes, inheriting the same cross-regime interference.
Despite these differences, existing approaches share a common limitation, namely regime-agnostic supervision.

\begin{figure}[htbp]
\centering
\includegraphics[width=0.7\linewidth]{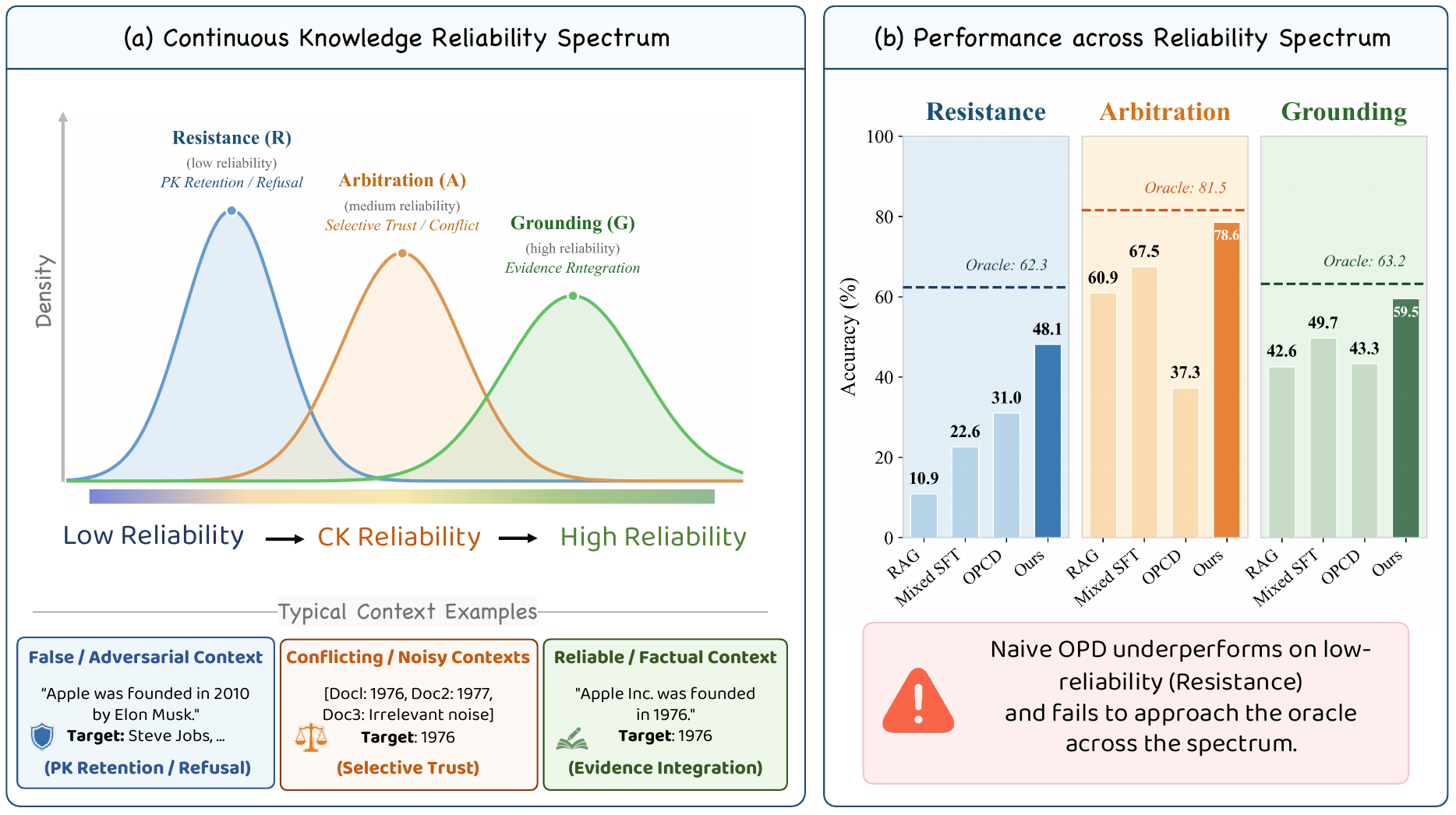}
\caption{Motivation for regime-aware peer specialization. \textbf{(a)} Contextual-knowledge reliability is modeled as a continuous spectrum and partitioned into three regimes---Resistance ($R$, low reliability), Arbitration ($A$, medium), and Grounding ($G$, high)---each requiring a distinct behavior: PK retention, selective trust, or evidence integration. \textbf{(b)} Per-regime accuracy of representative methods (bars) against the oracle peer specialist upper bound (dashed lines). RAG and Mixed SFT collapse on Resistance; single-teacher OPD (OPCD) improves over them but still falls well short of the oracle across all regimes. \ours{} substantially narrows this gap without requiring regime labels at inference.}
\label{fig:intro}
\end{figure}

Adapting OPD to knowledge conflict further exposes two coupled difficulties.
\textbf{Sample-level regime heterogeneity}. 
We distinguish three reliability regimes in RAG (Figure~\ref{fig:intro}a), namely \emph{Grounding} (reliable context to integrate), \emph{Arbitration} (mixed-quality context requiring selective trust), and \emph{Resistance} (adversarial context to reject), each demanding markedly different behaviors. 
Empirically, models trained per-regime outperform a jointly-trained model on their corresponding subsets, revealing a clear regime-wise specialization pattern. 
As a result, a regime-agnostic teacher receives conflicting optimization signals from different regimes and systematically underfits them all.
\textbf{Token-level supervision noise}. 
In conflict RAG, pivotal tokens often correspond to answer entities, source-attribution markers, rejection phrases, or option tokens where a small change determines whether the model follows CK or falls back to PK. 
The challenge is amplified in a multi-teacher setting, where disagreement among teachers introduces additional supervision heterogeneity~\citep{fu2026revisiting,tip2026,ko2026reopold}. 
Consequently, naively distilling from multiple specialized teachers does not reliably transfer their expertise to the student (Figure~\ref{fig:intro}b).

To address these challenges, we propose \textbf{\ours{}} (\textbf{R}egime-\textbf{A}ware \textbf{P}eer \textbf{S}pecialization with \textbf{D}ifficulty \textbf{A}nnealing), a training framework that handles heterogeneous knowledge conflicts through regime-aware peer supervision and difficulty-aware token learning.
All peer teachers are fine-tuned from a shared base model on regime-specific data, so the improvements come from specialization at a fixed model scale rather than from using a stronger or larger teacher.
The framework has three modules: regime-routed peer supervision, conflict-aware token selection, and difficulty annealing. 
The first two modules target sample-level regime heterogeneity and token-level supervision noise, respectively, while the third stabilizes their interaction during training.
\textbf{(i)~Regime-routed peer supervision} (\emph{addresses sample-level regime heterogeneity}). 
We train regime-specialized peer teachers and hard-route each sample to its matched peer for on-policy reverse-KL supervision, with the base model serving as an implicit KL anchor. 
Hard routing eliminates the contradictory gradient updates that a single jointly-trained teacher would impose across regimes.
\textbf{(ii)~Conflict-aware token selection} (\emph{addresses token-level supervision noise}). 
Three diagnostic signals (inter-teacher conflict, student--teacher gap, and student entropy) drive a hard mask that filters uninformative or unstable tokens, together with a soft weight that emphasizes tokens where the student is confidently misaligned with the regime-matched teacher.
\textbf{(iii)~Difficulty annealing} (stabilizes the combination of (i) and (ii)). 

Routing concentrates supervision on regime-matched teachers, while token selection further amplifies high-conflict positions.
Applying both mechanisms from the beginning can prematurely focus training on a small set of difficult tokens. 
We therefore start with broad token coverage and gradually concentrate supervision on harder positions as training progresses.
ALL specialization and routing occur only during training. 
At inference time, the deployed system is a single student model that requires neither regime labels nor access to peer teachers.

Our key observation is that sample-level routing and token-level selection address complementary challenges.
Routing resolves policy mismatch across reliability regimes, while token selection improves the quality of supervision within each regime.
Together, however, they can over-concentrate training on difficult tokens before the student has learned stable regime-specific behaviors.
Difficulty annealing mitigates this effect by gradually shifting supervision from broad coverage to high-conflict tokens.
Empirically, each module helps on its own, but only their combination yields the largest gains in our experiments.
Empirically, the three components are complementary, and their combination provide the most substantial benefits.

Our main contributions are as follows.
\begin{itemize}[leftmargin=*,nosep]
\item We characterize knowledge conflict in RAG as a spectrum of reliability regimes and show that regime-agnostic supervision induces conflicting optimization signals, motivating regime-specialized supervision.
\item We propose \ours{}, a regime-aware peer specialization framework that combines regime-routed peer supervision, conflict-aware token selection, and difficulty annealing to address heterogeneous knowledge conflicts at both sample and token levels.
\item Experiments across multiple knowledge-conflict benchmarks demonstrate that the proposed components are complementary and consistently outperform RL, single-teacher, and naive multi-teacher baselines, while generalizing to held-out conflict scenarios and benchmarks.
\end{itemize}

\section{Related Work}\label{sec:related}

We review related literature from two perspectives, namely robust RAG under knowledge conflict and on-policy distillation with token selection.

\subsection{Knowledge Conflict in RAG}\label{sec:rw-conflict}

RAG systems are vulnerable to noisy, outdated, or adversarial retrieved content, and the resulting knowledge conflicts manifest in diverse forms~\citep{xu2024knowledge,jin2024tug}.
Existing robustness methods span several families.
\emph{Prompting methods} guide models to verify or filter context before generation, including adaptive retrieval~\citep{ding2024retrieve}, conflict resolution prompting~\citep{wang2023resolving}, and chain-of-verification~\citep{he2024retrieving}.
\emph{Decoding methods} adjust the output distribution at inference time. CAD~\citep{shi2024cad} contrasts outputs with and without context, while DoLa~\citep{chuang2024dola} and DCCD~\citep{zhang2024dccd} contrast logits across layers or with dynamic strategies.
\emph{Training methods} directly optimize parameters. Self-RAG~\citep{asai2024self} learns to retrieve and self-critique through reflection tokens; InFO-RAG~\citep{xu2024inforag} aligns information flow for noise filtering; RAAT~\citep{fang2024raat} performs adversarial training against retrieval noise; KAFT~\citep{li2024kaft} fine-tunes on counterfactual contexts; Astute-RAG~\citep{wang2024astute} consolidates conflicting sources before answering.
\emph{RL-based methods} learn conflict-resolution policies through outcome-based optimization. Knowledgeable-R1~\citep{lin2025knowledgeable} applies GRPO~\citep{shao2024deepseekmath} with asymmetric advantage modulation, showing strong adversarial results.

These methods have substantially advanced RAG robustness. Prompting and decoding approaches provide lightweight, training-free conflict mitigation; supervised methods such as RAAT and KAFT equip models with parametric resilience to noisy or counterfactual context; and RL-based methods like Knowledgeable-R1 demonstrate that outcome-level optimization can yield strong adversarial robustness.
However, existing methods generally treat conflict as a single homogeneous phenomenon rather than explicitly distinguishing different reliability regimes.
Prompting and decoding approaches act at inference time without altering parameters; training and RL approaches adopt a single-policy paradigm that must reconcile opposing behaviors (e.g., following vs.\ rejecting context) within one model.
Moreover, outcome-level RL provides sparse credit assignment, offering limited signal for identifying which tokens caused errors.
Our work explores an alternative direction by partitioning conflict into three reliability regimes and assigning each to a same-scale peer specialist, with dense token-level on-policy supervision providing the training signal.

\subsection{On-Policy Distillation and Token Selection}\label{sec:rw-opd}

On-policy distillation (OPD) provides token-level teacher feedback on the student's own rollouts, combining distribution-matching with on-policy exploration~\citep{agarwal2024gkd,gu2024minillm}.
Recent advances include \emph{stability} improvements via top-$K$ truncated reverse-KL~\citep{fu2026revisiting} and length-inflation correction~\citep{luo2026stable}; \emph{token selection} strategies such as TIP's entropy-divergence taxonomy~\citep{tip2026}, SCOPE's correctness-based weighting~\citep{scope2026}, and SRPO's answer-correctness routing~\citep{srpo2026}; \emph{theoretical} results showing OPD equals dense KL-constrained RL (G-OPD; \citealp{yang2026gopd}) and establishing teacher consistency as necessary for stable training~\citep{wu2026lightning}; and \emph{self-distillation} extensions including OPCD~\citep{ye2026opcd} and privileged-information distillation~\citep{zhao2026opsd}.

In the multi-teacher setting, classic ensemble-then-distill~\citep{hinton2015distilling} averages teacher logits uniformly; BTX~\citep{sukhbaatar2024btx} trains specialized expert branches merged into an MoE deployed at inference; FuseLLM~\citep{wan2024fusellm} fuses heterogeneous LLM distributions; CA-MKD~\citep{camkd2024} selects teachers per instance based on confidence.

Two observations motivate our design.
First, recent OPD advances have made significant strides in stability and token-level credit assignment such as top-$K$ truncation~\citep{fu2026revisiting} tames heavy-tailed gradients, and TIP~\citep{tip2026} shows that profiling token importance yields meaningful gains. These works, however, typically assume a single teacher or use pre-defined domain labels (e.g., math vs.\ code) for multi-teacher routing; organizing teachers around conflict-derived reliability regimes within a single task remains unexplored.
Second, multi-teacher methods such as BTX and CA-MKD have demonstrated that instance-level teacher selection can outperform uniform ensembles. Yet existing token-selection strategies generally rely on one or two signals and apply a fixed selection policy throughout training, without adapting the difficulty of supervision as the student matures.
Building on these advances, \ours{} routes training samples to regime-matched peer specialists via hard assignment, applies a dual-layer token selector combining three complementary signals (inter-teacher conflict, student--teacher gap, and student entropy), and introduces difficulty annealing that progressively focuses the token budget on harder positions as training proceeds.

\section{Method}\label{sec:method}

\subsection{Overview}\label{sec:overview}

Given a question $x$ and a retrieved context $c$, a language model $\pi_\theta$ generates an answer $y \sim \pi_\theta(\cdot \mid x, c)$.
We categorize the reliability of $c$ relative to the model's parametric knowledge into three regimes $r \in \mathcal{R} = \{G, A, R\}$ along a single axis, \emph{how much could model trust context.}
\emph{Grounding}~($G$): the context is factually correct and relevant, and the model should integrate it faithfully.
\emph{Arbitration}~($A$): the context mixes reliable and unreliable evidence (including internally contradictory or irrelevant passages), and the model should selectively trust parts of it.
\emph{Resistance}~($R$): the context is adversarial or counterfactual, and the model should reject it and rely on parametric knowledge.
This trichotomy is defined by the \emph{logical relationship} between context and ground truth, independent of any particular benchmark. 
In Section~\ref{sec:modes} we show how existing conflict scenarios map onto these regimes.
Regime labels are used only during training to assign teachers, and the deployed student infers appropriate behavior from $(x, c)$ alone.
Our goal is to train a single student $\pi_\theta$ that handles all three regimes, using three same-scale peer teachers $\{\pi_{T_r}\}_{r \in \mathcal{R}}$ initialized from a shared base model $\pi_{\mathrm{base}}$.

\begin{figure}[t]
  \centering
  \includegraphics[width=0.95\linewidth]{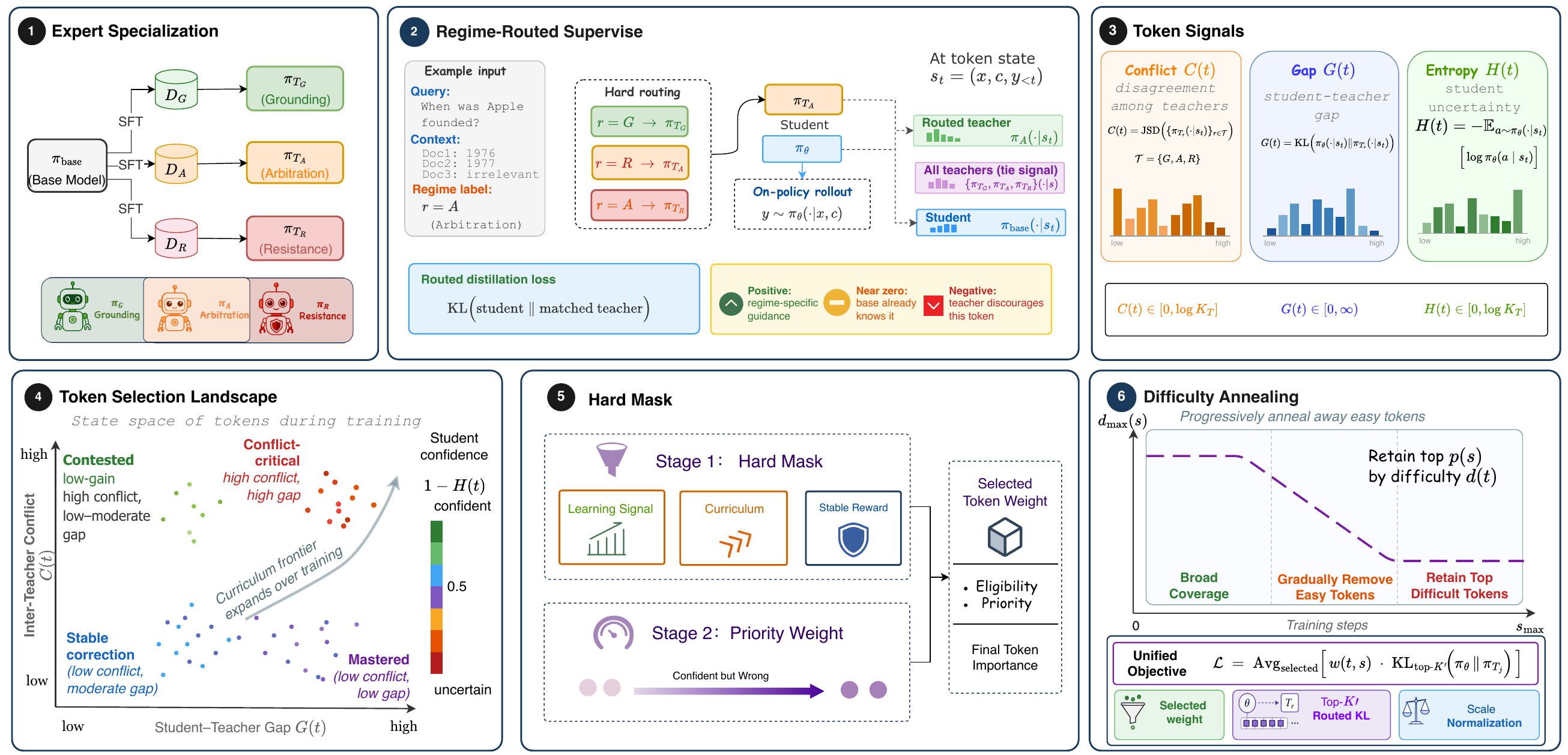}
  \caption{Overview of the \ours{} framework. \textbf{Panel~1:} A shared base model is fine-tuned into three regime-specialized peer teachers via SFT on the Grounding, Arbitration, and Resistance subsets. \textbf{Panel~2:} Each training sample is hard-routed by its regime label to the matched teacher; the student generates on-policy rollouts and receives routed reverse-KL supervision. \textbf{Panel~3:} Three token-level diagnostic signals---inter-teacher conflict~$\mathcal{C}(t)$, student\textendash teacher gap~$\mathcal{G}(t)$, and student entropy~$\mathcal{H}(t)$---are computed at each position. \textbf{Panel~4:} The token selection landscape, spanned by $\mathcal{G}(t)$ and $\mathcal{C}(t)$, partitions tokens into four regions; a curriculum frontier expands over training to cover increasingly difficult tokens. \textbf{Panel~5:} A dual-layer selector first applies a hard mask (information, curriculum, and stability filters) for eligibility, then a soft priority weight that upweights confident-but-misaligned tokens. \textbf{Panel~6:} Difficulty annealing progressively narrows the retained token set from broad coverage to the top-$p(s)$ most difficult tokens, and the unified objective aggregates the selected, weighted reverse-KL loss.}
  \label{fig:framework}
\end{figure}

To this end, \ours{} addresses the two challenges identified in Section~\ref{sec:intro} with two corresponding modules, plus a third that stabilizes their combination.
\textbf{(i)~Regime-routed peer supervision} (\S\ref{sec:routed-opd}). 
Because a single jointly-trained teacher imposes contradictory gradient updates across regimes, we train peer teachers specialized per regime and hard-route each sample to its matched peer for on-policy reverse-KL supervision.
\textbf{(ii)~Conflict-aware token selection} (\S\ref{sec:token-selection}--\ref{sec:soft-weight}). 
Pivotal tokens carry disproportionate influence yet are the most susceptible to inter-teacher disagreement and student uncertainty. 
% A dual-layer selector uses 
We thus introduce a dual-layer selector based on three diagnostic signals: inter-teacher conflict~($\mathcal{C}$), student--teacher gap~($\mathcal{G}$), and student entropy~($\mathcal{H}$).
The hard mask filters uninformative or unstable tokens, while the soft weight upweights confidently misaligned ones.
\textbf{(iii)~Difficulty annealing} (\S\ref{sec:annealing}). 
In early training, the student needs broad distributional coverage to build regime-consistent foundations. 
Restricting supervision to high-conflict tokens too early can hinder this process. 
An annealing schedule starts with the full token set and progressively focuses on the most difficult tokens as the student matures.
Figure~\ref{fig:framework} provides an overview of the complete pipeline.

\subsection{Regime-Routed Peer Supervision}\label{sec:routed-opd}

\paragraph{Regime-specialized peer teachers.}
We partition the training data by the three regimes defined above and train three same-scale peer teachers from $\pi_{\mathrm{base}}$ via supervised fine-tuning on each subset $\mathcal{D}_r$:
\begin{equation}\label{eq:teacher-sft}
  \pi_{T_r} \!=\! \argmin_{\pi}\; \E_{(x,c,y^*) \sim \mathcal{D}_r} \bigl[-\log \pi(y^* \mid x, c)\bigr],
\end{equation}
where $y^*$ is the regime-appropriate gold response (Section~\ref{sec:modes}).
Training a single teacher across all regimes leads to cross-regime interference (Section~\ref{sec:ablation}). Same-scale per-regime teachers isolate specialization from capacity scaling.

\paragraph{On-policy reverse-KL distillation.}
Each training sample is hard-routed to its regime-matched teacher.
Unlike standard distillation that minimizes forward KL on gold trajectories, our goal is \emph{on-policy correction}: the student generates from its own policy and the teacher corrects behavior under the prefixes the student will actually produce at deployment.
We accordingly use on-policy rollouts so that the student is exposed to its own mistakes, and adopt reverse KL to encourage mode-seeking behavior that concentrates mass on tokens the routed teacher considers likely while suppressing those it deems unlikely.
The per-sample on-policy reverse-KL objective is
\begin{equation}\label{eq:per-sample-loss}
  \mathcal{L}_r(\theta) = \E_{y \sim \pi_\theta(\cdot|x,c)} \left[\frac{1}{|y|}\sum_{t=1}^{|y|} D_{\mathrm{KL}}\!\left(\pi_\theta(\cdot \mid \mathbf{s}_t) \;\big\|\; \pi_{T_r}(\cdot \mid \mathbf{s}_t)\right)\right],
\end{equation}
where $\mathbf{s}_t = (x, c, y_{<t})$ denotes the prefix state at position $t$.

\paragraph{From distribution KL to token-level implicit reward.}
To expose the per-token learning signal hidden inside Eq.~\eqref{eq:per-sample-loss}, we follow the G-OPD framework~\citep{yang2026gopd} and rewrite the reverse KL by introducing the base model $\pi_{\mathrm{base}}$ as a reference distribution.
Expanding the KL definition and adding-and-subtracting $\log \pi_{\mathrm{base}}$ inside the logarithm gives
\begin{align}
  D_{\mathrm{KL}}(\pi_\theta \| \pi_{T_r})
  &= \E_{\pi_\theta}\!\left[\log \frac{\pi_\theta}{\pi_{T_r}}\right] \notag\\
  &= \E_{\pi_\theta}\!\left[\log \frac{\pi_\theta}{\pi_{\mathrm{base}}}\right] - \E_{\pi_\theta}\!\left[\log \frac{\pi_{T_r}}{\pi_{\mathrm{base}}}\right] \notag\\
  &= D_{\mathrm{KL}}(\pi_\theta \| \pi_{\mathrm{base}}) - \E_{\pi_\theta}\!\left[r_t^{(r)}\right],\label{eq:gopd-derivation}
\end{align}
where each sampled token $a_t$ receives a dense \emph{implicit reward}
\begin{equation}\label{eq:implicit-reward}
  r_t^{(r)} = \log \frac{\pi_{T_r}(a_t \mid \mathbf{s}_t)}{\pi_{\mathrm{base}}(a_t \mid \mathbf{s}_t)}.
\end{equation}
Substituting Eq.~\eqref{eq:gopd-derivation} into the per-sample loss yields the equivalent KL-constrained reward maximization
\begin{equation}\label{eq:gopd-equiv}
  \min_\theta\; \mathcal{L}_r(\theta) \;\Longleftrightarrow\; \max_\theta \left\{ \E_{\pi_\theta}\!\left[r_t^{(r)}\right] - D_{\mathrm{KL}}(\pi_\theta \| \pi_{\mathrm{base}}) \right\}.
\end{equation}
Two properties follow.
(i) $\pi_{\mathrm{base}}$ serves as an \emph{implicit KL anchor}, providing regularization without an explicit reference term in the loss.
(ii) Each token carries a dense, regime-specific learning signal $r_t^{(r)}$ that reflects how much the routed teacher agrees with the base at that position. Unlike RL's sparse outcome-level reward, this dense signal enables the dual-layer token selector in Section~\ref{sec:token-selection} to operate on a principled reward rather than on raw KL contributions.
The full training objective aggregates Eq.~\eqref{eq:per-sample-loss} across regimes via hard routing, so that each sample activates exactly one teacher with no distribution mixing.

\subsection{Token-Level Dual-Layer Selection}\label{sec:token-selection}

While regime routing assigns an appropriate teacher to each sample, supervision quality can still vary substantially across tokens.
Some tokens provide little learning signal because the student is already aligned with the teacher~\citep{tip2026,ko2026reopold}.
Others produce high-variance gradients that destabilize training~\citep{ko2026reopold}. 
Still others lie in high-conflict regions where teachers disagree.
We therefore separate token selection into two stages: a \emph{hard eligibility mask} $M(t)$ that removes tokens likely to inject harmful or premature supervision, and a \emph{soft weight} $\omega(t, s)$ that assigns graded emphasis among the surviving tokens.

\subsubsection{Token-Level Supervision Signals}

No single signal can reliably distinguish informative tokens from uninformative or destabilizing ones, because inter-teacher consistency, student--teacher divergence, and student confidence 
jointly affect whether a token provides useful supervision.
% interact in non-trivial ways.
We therefore introduce three complementary diagnostic signals, all normalized to $[0,1]$ before use.

\paragraph{Inter-teacher conflict $\mathcal{C}(t)$.}
This signal measures how much the $K$ peer teachers disagree at a given token, defined as the Jensen--Shannon divergence:
\begin{equation}\label{eq:conflict}
  \mathcal{C}(t) = H\!\left(\frac{1}{K}\sum_{k=1}^{K} \pi_{T_k}(\cdot|\mathbf{s}_t)\right) - \frac{1}{K}\sum_{k=1}^{K} H\!\left(\pi_{T_k}(\cdot|\mathbf{s}_t)\right),
\end{equation}
where $H(\cdot)$ denotes entropy and the mixture is over all $K=3$ teachers.
$\mathcal{C}(t) = 0$ when all teachers agree; large $\mathcal{C}(t)$ indicates a token at the core of knowledge conflict.
Note that while supervision comes exclusively from the routed teacher, $\mathcal{C}(t)$ queries all teachers to assess disagreement.
For efficiency, $\mathcal{C}(t)$ is approximated on the union of each teacher's top-$K'$ tokens and refreshed every $N$ training steps.

\paragraph{Student--teacher gap $\mathcal{G}(t)$.}
This signal measures how far the student deviates from the routed teacher at each position, defined as the per-token reverse-KL:
\begin{equation}\label{eq:gap}
  \mathcal{G}(t) = D_{\mathrm{KL}}\!\left(\pi_\theta(\cdot|\mathbf{s}_t) \;\big\|\; \pi_{T_r}(\cdot|\mathbf{s}_t)\right),
\end{equation}
$\mathcal{G}(t) \approx 0$ means the student has already aligned with the teacher at this position.
Large $\mathcal{G}(t)$ indicates that corrective supervision is still needed.
% Since $\mathcal{G}(t)$ is the per-token training loss, it requires no additional computation.

\paragraph{Student entropy $\mathcal{H}(t)$.}
Inspired by the ``confident-but-misaligned'' blind spot identified by~\citet{tip2026}, we introduce a normalised entropy signal to capture the student's predictive uncertainty at position $t$:
\begin{equation}\label{eq:entropy}
  \mathcal{H}(t) = \frac{H\!\left(\pi_\theta(\cdot|\mathbf{s}_t)\right)}{\log |\mathcal{V}|} \in [0, 1],
\end{equation}
where $|\mathcal{V}|$ is the vocabulary size.
Low $\mathcal{H}(t)$ indicates a confident student, which is desirable when correct but dangerous when misaligned with the teacher.

\subsubsection{Hard Mask: Information and Stability Filters}\label{sec:hard-mask}

Not all tokens within a regime-routed sample provide useful supervision.
% Two distinct risks motivate a hard eligibility mask. 
We use a hard eligibility mask to control two risks.
First, tokens where the student has already aligned with the teacher carry negligible gradient signal and waste computation. 
Second, tokens with extreme negative implicit rewards under high student uncertainty arise from off-manifold rollouts where teacher supervision is unreliable and can destabilize gradients.
A token is eligible only if both conditions hold:
\begin{equation}\label{eq:hard-mask}
M(t) = \mathbb{1}\!\left[\mathcal{G}(t) \geq \epsilon_{\mathcal{G}}(s)\right] \cdot \mathbb{1}\!\left[\neg\bigl(r_t^{(r)} < r_{\min} \;\land\; \mathcal{H}(t) > h_{\max}\bigr)\right],
\end{equation}
where the first factor is the \emph{information filter} and the second is the \emph{stability filter}, described below.

\paragraph{Information filter.}
Tokens with $\mathcal{G}(t) < \epsilon_{\mathcal{G}}(s)$ have already been learned and carry negligible gradient signal.
The threshold is set as an adaptive percentile computed per regime within each batch, since gap distributions differ substantially across regimes (e.g., Grounding tokens tend to have smaller gaps than Resistance tokens):
\[
  \epsilon_{\mathcal{G}}(s) = Q_{\beta}\!\left(\{\mathcal{G}(t) : r_i = r\}_{t \in \mathrm{batch}}\right),
\]
where $Q_{\beta}(\cdot)$ denotes the $\beta$-th quantile.
This filters the bottom $\beta$\% of tokens by learning gap within each regime in each batch, preventing systematic over-filtering of regimes with inherently smaller gaps while adapting automatically as the student improves..
% The adaptive percentile automatically adjusts as the student improves over training.

\paragraph{Stability filter.}
Not all tokens with negative implicit rewards should be discarded.
A token where the student is confident yet strongly disfavored by the regime teacher ($r_t^{(r)} \ll 0$, low $\mathcal{H}$) is precisely a confident-but-misaligned prediction that the soft weight should later prioritize for correction.
Removing it would eliminate the most valuable corrective signal.
We therefore filter only tokens that are both extremely disfavored and associated with high student uncertainty ($\mathcal{H}(t) > h_{\max}$), which typically arise from off-manifold student rollouts where teacher supervision is unreliable.
Following the mixture-based lower-bound derivation of \citet{ko2026reopold}, the reward threshold is set as $r_{\min} = \log \lambda_{\mathrm{clip}} / (1 - \lambda_{\mathrm{clip}})$, where $\lambda_{\mathrm{clip}}$ is the mixture coefficient that controls the clipping aggressiveness~\citealp{ko2026reopold}, $h_{\max}$ is a student-entropy ceiling.
When the student is confident (low $\mathcal{H}$), the stability condition is automatically satisfied regardless of the reward, ensuring that confident mismatches are retained for correction.

\subsubsection{Soft Weight for Correction Prioritisation}\label{sec:soft-weight}

Among the tokens that pass the hard mask, not all surviving tokens deserve equal attention.
The most valuable corrective signal comes from tokens where the student is confident yet far from the routed teacher, i.e., the ``confident-but-misaligned'' blind spot identified by~\citet{tip2026}.
These tokens represent cases where the student has committed to an incorrect direction and ordinary gradient updates may be insufficient to reverse the commitment.
In contrast, tokens where the student is already uncertain or close to the teacher need only standard-weight supervision.

We therefore introduce a soft weight that assigns graded priority among eligible tokens:
\begin{equation}\label{eq:soft-weight}
  \omega(t, s) = 1 + \eta \cdot (1 - \mathcal{H}(t)) \cdot \mathcal{G}(t),
\end{equation}
where $\eta$ controls the boost magnitude.
When the student is uncertain, i.e.\ $\mathcal{H}(t)$ is large, or already close to the teacher, i.e.\ $\mathcal{G}(t)$ is small, the weight reduces to $\omega \approx 1$ and standard supervision applies.

The hard mask eliminates tokens that would waste computation or destabilize training, while the soft weight fine-tunes priority among the remaining informative tokens.

\subsection{Difficulty Annealing}\label{sec:annealing}

Routing and token selection together concentrate supervision on high-signal, high-conflict tokens.
In early training, however, the student has not yet acquired regime-consistent foundations, so restricting supervision to only high-conflict tokens from the start would deprive it of the broad distributional coverage needed to build basic competence.
A natural remedy is to begin with the full token set and then progressively anneal away easy tokens, gradually focusing the training budget on the most informative, high-conflict positions as the student matures.

The difficulty score combines inter-teacher conflict with a confidence interaction term:
\begin{equation}\label{eq:difficulty}
  d(t) = \mathcal{C}(t) + \gamma \cdot \mathcal{C}(t) \cdot (1 - \mathcal{H}(t)),
\end{equation}
where $\gamma$ controls the interaction strength.
The first term captures raw conflict intensity.
The interaction term adds difficulty when the student is both in a high-conflict region and highly confident there, identifying tokens where the student may have committed to an incorrect direction.
The annealing schedule controls how many tokens survive via a retention ratio $p(s)$ that decreases linearly from~1 to a final value $p_{\mathrm{final}}$ over an annealing horizon $S_{\mathrm{a}}$, and remains constant thereafter:
\begin{equation}\label{eq:annealing-schedule}
  p(s) =
  \begin{cases}
    1 - (1 - p_{\mathrm{final}})\,\dfrac{s}{S_{\mathrm{a}}}, & s < S_{\mathrm{a}}, \\[4pt]
    p_{\mathrm{final}}, & s \geq S_{\mathrm{a}}.
  \end{cases}
\end{equation}
A token is retained by the annealing mask if its difficulty ranks in the top $p(s)$ fraction within the sequence:
\begin{equation}\label{eq:annealing-mask}
  A(t, s) = \mathbb{1}\!\left[d(t) \geq Q_{1-p(s)}\!\bigl(\{d(t')\}_{t'}\bigr)\right].
\end{equation}
At the start of training $p(s){=}1$ and all tokens pass the annealing filter.
By step $S_{\mathrm{a}}$, only the top $p_{\mathrm{final}}$ fraction of difficult tokens is retained.
This transitions from broad coverage to focused, high-conflict supervision at the token level, rather than a coarser regime-level ordering.

Note that the annealing and soft weighting interact by design.
The annealing gradually removes easy tokens whose learning signal has diminished, while the soft weight prioritizes the surviving high-conflict, confident-but-misaligned tokens for correction.
The annealing parameters ($p_{\mathrm{final}}$, $S_{\mathrm{a}}$) are set once and shared across all regimes.

\subsection{Unified Training Objective}\label{sec:unified-objective}

Combining all components, we first define the effective per-token weight:
\begin{equation}\label{eq:effective-weight}
  w(t, s) = M(t) \cdot A(t, s) \cdot \omega(t, s),
\end{equation}
which folds together the hard mask, annealing mask and the soft correction-priority weight.
The full \ours{} training objective is then
\begin{equation}\label{eq:final-loss}
\begin{split}
\mathcal{L}(\theta) &= \E_{(x,c,r) \sim \mathcal{D}} \; \E_{y \sim \pi_\theta} \\
&\quad \left[ \frac{1}{Z} \sum_{t=1}^{|y|} w(t, s) \cdot D_{\mathrm{KL}}^{K'}\!\left(\pi_\theta(\cdot|\mathbf{s}_t) \;\big\|\; \pi_{T_r}(\cdot|\mathbf{s}_t)\right) \right],
\end{split}
\end{equation}
where $Z = \max\!\left(\sum_t w(t,s),\; Z_{\min}\right)$ normalises by the effective token count and is clipped from below to avoid high-variance gradients when few tokens survive.
$D_{\mathrm{KL}}^{K'}$ denotes the top-$K'$ truncated reverse KL~\citep{fu2026revisiting}, which restricts the computation to the union of the teacher's and student's top-$K'$ tokens with renormalization, preserving the reverse-KL penalty on out-of-support student assignments via a residual bucket.
Algorithm~\ref{alg:raps-da} summarizes the complete training procedure.

\begin{algorithm}[t]
\caption{\ours{} training (one step).}
\label{alg:raps-da}
\begin{algorithmic}[1]
\Statex \textbf{Input:} student $\pi_\theta$, regime teachers $\{\pi_{T_r}\}$, base $\pi_{\mathrm{base}}$, batch $\mathcal{B}$ with regime labels, step $s$.
\Statex
\Statex \texttt{// Stage 1: Annealing schedule}
\State Compute retention ratio $p(s)$ from annealing schedule.
\Statex
\Statex \texttt{// Stage 2: On-policy rollout and signal computation}
\For{each $(x_i, c_i, r_i) \in \mathcal{B}$}
    \State Generate rollout $y_i \sim \pi_\theta(\cdot \mid x_i, c_i)$.
    \State Compute per-token signals $\mathcal{G}(t)$, $\mathcal{H}(t)$, and $r_t^{(r)}$.
    \State Refresh inter-teacher conflict $\mathcal{C}(t)$ every $N$ steps.
\EndFor
\Statex
\Statex \texttt{// Stage 3: Token selection}
\State Batch-normalize $\mathcal{C}$, $\mathcal{G}$ to $[0,1]$.
\State Form hard mask $M(t)$ via information and stability filters.
\State Compute difficulty $d(t)$ and retain top-$p(s)$ fraction as annealing mask $A(t, s)$.
\State Form soft weight $\omega(t,s) \gets 1 + \eta \cdot (1 - \mathcal{H}(t)) \cdot \mathcal{G}(t)$.
\Statex
\Statex \texttt{// Stage 4: Gradient update}
\State Compute $\mathcal{L}(\theta)$ with effective weight $w(t,s) = M(t) \cdot A(t,s) \cdot \omega(t,s)$.
\State Update $\theta \gets \theta - \mathrm{lr} \cdot \nabla_\theta \mathcal{L}(\theta)$.
\end{algorithmic}
\end{algorithm}

\section{Experiments}\label{sec:experiments}

We evaluate \ours{} across multiple knowledge-conflict scenarios, examining whether regime-aware peer specialization yields a single student that appropriately leverages or disregards retrieved context depending on its reliability, without relying on a stronger or larger teacher.

\subsection{Experimental Setup}

\subsubsection{Datasets}\label{sec:modes}

\paragraph{In-distribution.}
We adopt the five knowledge-conflict scenarios from the Knowledgeable-R1 (KR1) benchmark~\cite{lin2025knowledgeable}, each requiring a distinct behavior from the model:
S1 (correct context) provides accurate supporting evidence that should be faithfully integrated;
S2 (adversarial context) contains deliberately counterfactual passages that should be rejected;
S3 (self-conflicting context) presents internally contradictory claims requiring arbitration;
S4 (irrelevant context) includes topically unrelated passages that should be ignored;
and S5 (partially relevant context) mixes valid evidence with distractors, demanding selective extraction.
% Each sample includes top-5 retrieved paragraphs following the KR1 protocol.
For our three-regime formulation ($\mathcal{R}=\{G,A,R\}$), we group these scenarios by the dominant behavior required.
S1 and S5 map to Grounding ($G$), where the model should leverage reliable or partially relevant context.
S3 and S4 map to Arbitration ($A$), where the model must arbitrate among conflicting or irrelevant evidence.
S2 maps to Resistance ($R$), where the model must reject misleading evidence.

\paragraph{Out-of-distribution.}
To test whether the learned conflict-resolution policy generalizes beyond the training distribution, we evaluate on two held-out benchmarks not used in any training stage:
\textbf{(1)~ConflictQA}~\cite{su2024conflictbank} (PopQA subset, 7{,}947 samples) pairs each factual question with a parametric-consistent (PC) instance whose context supports the correct answer and a non-consistent (NC) instance whose context contains counterfactual evidence. Comparing PC vs.\ NC accuracy directly measures conflict robustness.
\textbf{(2)~TruthfulQA}~\cite{lin2022truthfulqa} (817 questions) probes whether the model retains truthful parametric knowledge. We evaluate in a no-context setting without retrieved passages to test whether training preserves the base model's parametric accuracy.

\subsubsection{Evaluation Metrics}\label{sec:eval-metrics}

We adopt exact match (EM) as the primary metric, consistent with prior work~\cite{jin2025search, lin2025knowledgeable}. For each scenario, EM is computed on the respective test or validation split; we additionally report unweighted regime-level averages ($G$/$A$/$R$) and overall averages.
For the OOD benchmarks, we report EM on ConflictQA under the RAG-augmented setting and EM on TruthfulQA under the no-context setting (testing parametric-knowledge preservation).

\subsubsection{Baselines}\label{sec:baselines}

We compare against the following methods, spanning prompting, decoding, fine-tuning, and distillation approaches:

\paragraph{Inference-time methods.}
\textbf{Query-only prompting} uses only parametric knowledge without retrieved context, serving as a lower bound for context-dependent scenarios and an upper bound when context is misleading.
\textbf{RAG prompting} prepends the top-5 retrieved passages to the query without additional training.
\textbf{Astute-RAG}~\cite{wang2024astute} consolidates and clusters conflicting sources before answering.
\textbf{CK-PLUG}~\cite{bi2025parametersvscontextfinegrained} is a plug-and-play decoding strategy that adjusts token probabilities at detected conflict spans.

\paragraph{Training-based methods.}
\textbf{Mixed SFT} performs supervised fine-tuning on the union of all regime-specific data, using the same total number of training samples as \ours{}.

\paragraph{RL methods.}
\textbf{GRPO w/ RAG}~\cite{guo2025deepseekr1_nature} applies group relative policy optimization with outcome-level reward on RAG rollouts.
\textbf{Knowledgeable-R1}~\cite{lin2025knowledgeable} extends GRPO with joint PK/CK/RPK sampling and adaptive advantage modulation, representing the state-of-the-art on the benchmark.
\paragraph{OPD methods.}
\textbf{OPCD}~\cite{ye2026opcd} conditions the teacher on context while the student generates without it, transferring contextual reasoning into parametric knowledge. Since OPCD uses a single jointly-trained teacher with the same OPD hyperparameters as \ours{}, it also serves as our single-teacher OPD baseline in the ablation study.
\textbf{TIP}~\cite{tip2026} profiles token importance and up-weights informative tokens via a learned scorer.

\subsubsection{Implementation Details}

We adopt \texttt{Qwen2.5-7B-Instruct}\footnote{\url{https://huggingface.co/Qwen/Qwen2.5-7B-Instruct}}~\cite{qwen2025qwen25technicalreport} as the shared base model for both the student and all peer teachers. This same-scale (7B$\to$7B) setting isolates the effect of regime-aware specialization from any teacher capacity advantage. All three peer teachers are initialized from the same checkpoint and fine-tuned independently on their regime-specific subsets (Section~\ref{sec:routed-opd}). For the Llama baseline in Table~\ref{tab:main-results}, we use \texttt{Llama3.1-8B-Instruct} from \url{https://huggingface.co/meta-llama/Llama-3.1-8B-Instruct}.

For OPD training, we use AdamW with a learning rate of $1 \times 10^{-6}$ and linear warmup over the first 10\% of steps. Each batch consists of 64 prompts with 4 rollouts each via top-$p$ sampling ($p{=}0.9$, $\tau{=}1.0$). The token-level reverse KL uses the teacher's top-$K'{=}32$ tokens with renormalization~\cite{fu2026revisiting}: concretely, let $\mathcal{S}_t = \mathrm{TopK}'(\pi_\theta(\cdot|\mathbf{s}_t)) \cup \mathrm{TopK}'(\pi_{T_r}(\cdot|\mathbf{s}_t))$ and define $\tilde{\pi}(v) = \pi(v)$ for $v \in \mathcal{S}_t$ and $\tilde{\pi}(\textsc{rest}) = \sum_{u \notin \mathcal{S}_t} \pi(u)$; the KL is computed over $\mathcal{S}_t \cup \{\textsc{rest}\}$. The inter-teacher conflict signal $\mathcal{C}(t)$ is refreshed every $N{=}50$ training steps; intermediate steps reuse the cached estimate. Special tokens are excluded from all signal computations and from the loss. The base model $\pi_{\mathrm{base}}$ serves as the implicit KL anchor. Training proceeds for 2 epochs on 8 NVIDIA H100 80GB GPUs in 2 hours. All results are averaged over three random schedules.

\subsection{Main Results}\label{sec:main-results}

Table~\ref{tab:main-results} reports per-scenario exact-match accuracy and regime-level averages across both backbone models.
We compare \ours{} against prompting, decoding, fine-tuning, and RL baselines to assess whether regime-aware peer specialization yields consistent improvements, and include the oracle peer specialist as a non-deployable upper bound that requires ground-truth regime labels at inference.

\begin{table}[t]
\centering
\setlength{\tabcolsep}{2.8pt}
\renewcommand{\arraystretch}{1.12}
\caption{Main in-distribution results (EM, \%) following the KR1 protocol~\cite{lin2025knowledgeable}. Best in \textbf{bold}, second-best \underline{underlined}; Oracle PS is excluded from ranking. \colorbox{teal!12}{Shaded rows} are our methods.}
\label{tab:main-results}
\begin{threeparttable}
\footnotesize
\resizebox{\linewidth}{!}{%
\begin{tabular}{l ccc ccc c c ccc | ccc}
\toprule
& \multicolumn{3}{c}{\textbf{S1 Correct ($G$)}} & \multicolumn{3}{c}{\textbf{S2 Wrong ($R$)}} & \textbf{S3 ($A$)} & \textbf{S4 ($A$)} & \multicolumn{3}{c}{\textbf{S5 Partly-Irr. ($G$)}} & \multicolumn{3}{c}{\textbf{Regime Avg.}}\\
\cmidrule(lr){2-4}\cmidrule(lr){5-7}\cmidrule(lr){8-8}\cmidrule(lr){9-9}\cmidrule(lr){10-12}\cmidrule(lr){13-15}
\textbf{Method} & PC-MR & PC-MC & PC-QA & NC-MR & NC-MC & NC-QA & SC & ExpPE & HotPot & 2Wiki & Musique & $G$ & $A$ & $R$\\
\midrule
\multicolumn{15}{l}{\texttt{Qwen2.5-7B-Instruct}} \\
Query-only        & 27.7 & 24.7 & 31.7 & 25.9 & 25.8 & 32.3 & 29.7 & 64.5 & 20.9 & 25.5 & 4.4  & 22.5 & 47.1 & 28.0 \\
RAG prompting     & 65.7 & 66.4 & 74.3 & 13.5 & 8.1  & 11.3 & 59.5 & 62.2 & 20.4 & 22.5 & 6.4  & 42.6 & 60.9 & 10.9 \\
CK-PLUG~\cite{bi2025parametersvscontextfinegrained} & 64.7 & 66.5 & 78.7 & 11.6 & 8.1 & 7.9 & 55.0 & 55.0 & 22.7 & 24.8 & 6.2 & 43.9 & 55.0 & 9.2 \\
Astute~\cite{wang2024astute}    & 65.5 & 66.0 & 77.6 & 12.8 & 7.1  & 10.3 & 54.2 & 56.7 & 17.9 & 20.4 & 6.3  & 42.3 & 55.5 & 10.1 \\
SFT               & 72.0 & 77.7 & 74.7 & 24.9 & 21.1 & 22.0 & 68.5 & 66.6 & 30.1 & 32.2 & 11.8 & 49.7 & 67.5 & 22.6 \\
GRPO w/ RAG~\cite{guo2025deepseekr1_nature} & 77.6 & 77.4 & \underline{80.0} & 26.9 & 19.7 & 26.0 & 75.3 & 66.5 & 27.9 & 34.0 & 11.8 & 51.4 & 70.9 & 24.2 \\
KR1~\cite{lin2025knowledgeable} & 75.1 & 75.5 & \textbf{80.9} & \underline{43.9} & \underline{37.3} & 29.4 & \underline{76.3} & \underline{67.6} & 31.4 & 37.5 & 12.0 & 52.1 & \underline{72.0} & 36.9 \\
OPCD~\cite{ye2026opcd} & 64.7 & 64.4 & 62.8 & 40.2 & 37.0 & 15.8 & 56.5 & 18.1 & 24.6 & 34.5 & 8.8 & 43.3 & 37.3 & 31.0 \\
TIP~\cite{tip2026} & \underline{81.8} & \underline{79.1} & 76.8 & 33.3 & 36.5 & \underline{41.0} & 76.2 & 52.9 & \underline{31.7} & \underline{44.5} & \underline{15.2} & \underline{54.9} & 64.5 & \underline{36.9} \\
\rowcolor{teal!12}
\textbf{\ours{} (ours)} & \textbf{89.3} & \textbf{87.0} & \textbf{80.9} & \textbf{51.5} & \textbf{48.2} & \textbf{44.6} & \textbf{87.3} & \textbf{69.9} & \textbf{33.9} & \textbf{45.5} & \textbf{20.1} & \textbf{59.5} & \textbf{78.6} & \textbf{48.1} \\
% \rowcolor{teal!4}
% Oracle PS \emph{(upper bd.)} & 94.9 & 92.9 & 92.1 & 74.1 & 64.1 & 48.6 & 92.2 & 70.8 & 34.4 & 48.4 & 16.5 & 63.2 & 81.5 & 62.3 \\
\midrule
\multicolumn{15}{l}{\texttt{Llama3.1-8B-Instruct}} \\
Query-only        & 29.4 & 26.2 & 39.9 & 27.1 & 27.6 & 42.6 & 32.1 & 43.3 & 20.7 & 21.0 & 6.2  & 23.9 & 37.7 & 32.5 \\
RAG prompting     & 64.8 & 62.0 & 76.4 & 22.9 & 16.3 & 24.9 & 61.2 & 39.2 & 24.4 & 23.5 & 8.2  & 43.2 & 50.2 & 21.4 \\
CK-PLUG           & 54.8 & 58.5 & 69.7 & 12.1 & 9.1  & 17.3 & 42.0 & 31.5 & 22.4 & 24.6 & 5.2  & 39.2 & 36.8 & 12.8 \\
Astute            & 65.8 & 64.9 & 78.0 & 17.0 & 9.1  & 17.3 & 59.8 & 40.1 & 1.6  & 30.3 & 9.6  & 41.7 & 50.0 & 14.4 \\
SFT               & 72.9 & 79.2 & 73.8 & 42.6 & 35.5 & 35.9 & 70.1 & 47.2 & 33.6 & 38.2 & 13.4 & 51.9 & 58.6 & 38.0 \\
GRPO w/ RAG       & \underline{78.0} & 79.7 & \underline{82.6} & 41.6 & 35.7 & 39.3 & \underline{76.6} & 47.6 & 34.8 & 41.2 & \underline{16.6} & \underline{55.5} & \underline{62.1} & 38.8 \\
KR1               & 73.8 & \underline{80.2} & 80.0 & 55.4 & 41.1 & \underline{44.6} & 73.7 & \underline{49.6} & \underline{37.1} & \underline{45.4} & 14.7 & 55.2 & 61.6 & 47.0 \\
OPCD~\cite{ye2026opcd} & 61.8 & 59.5 & 70.2 & 46.7 & 40.4 & 32.9 & 54.2 & 25.6 & 27.8 & 38.7 & 8.5 & 44.4 & 39.9 & 40.0 \\
TIP~\cite{tip2026} & 75.4 & 78.1 & 45.8 & \underline{62.5} & \textbf{58.9} & 41.3 & 65.5 & 36.4 & 35.2 & 43.7 & 15.2 & 48.9 & 51.0 & \underline{54.2} \\
\rowcolor{teal!12}
\textbf{\ours{} (ours)} & \textbf{79.5} & \textbf{83.7} & \textbf{84.4} & \textbf{67.1} & \underline{52.9} & \textbf{55.7} & \textbf{77.8} & \textbf{52.6} & \textbf{40.2} & \textbf{49.5} & \textbf{18.3} & \textbf{59.2} & \textbf{65.2} & \textbf{58.6} \\
% \rowcolor{teal!4}
% Oracle PS \emph{(upper bd.)} & 93.1 & 94.3 & 90.6 & 79.8 & 67.3 & 61.4 & 91.2 & 53.8 & 42.4 & 55.9 & 20.1 & 66.1 & 72.5 & 69.5 \\
\bottomrule
\end{tabular}}
\end{threeparttable}
\end{table}
  
\begin{figure}[t]
\centering
\includegraphics[width=0.5\linewidth]{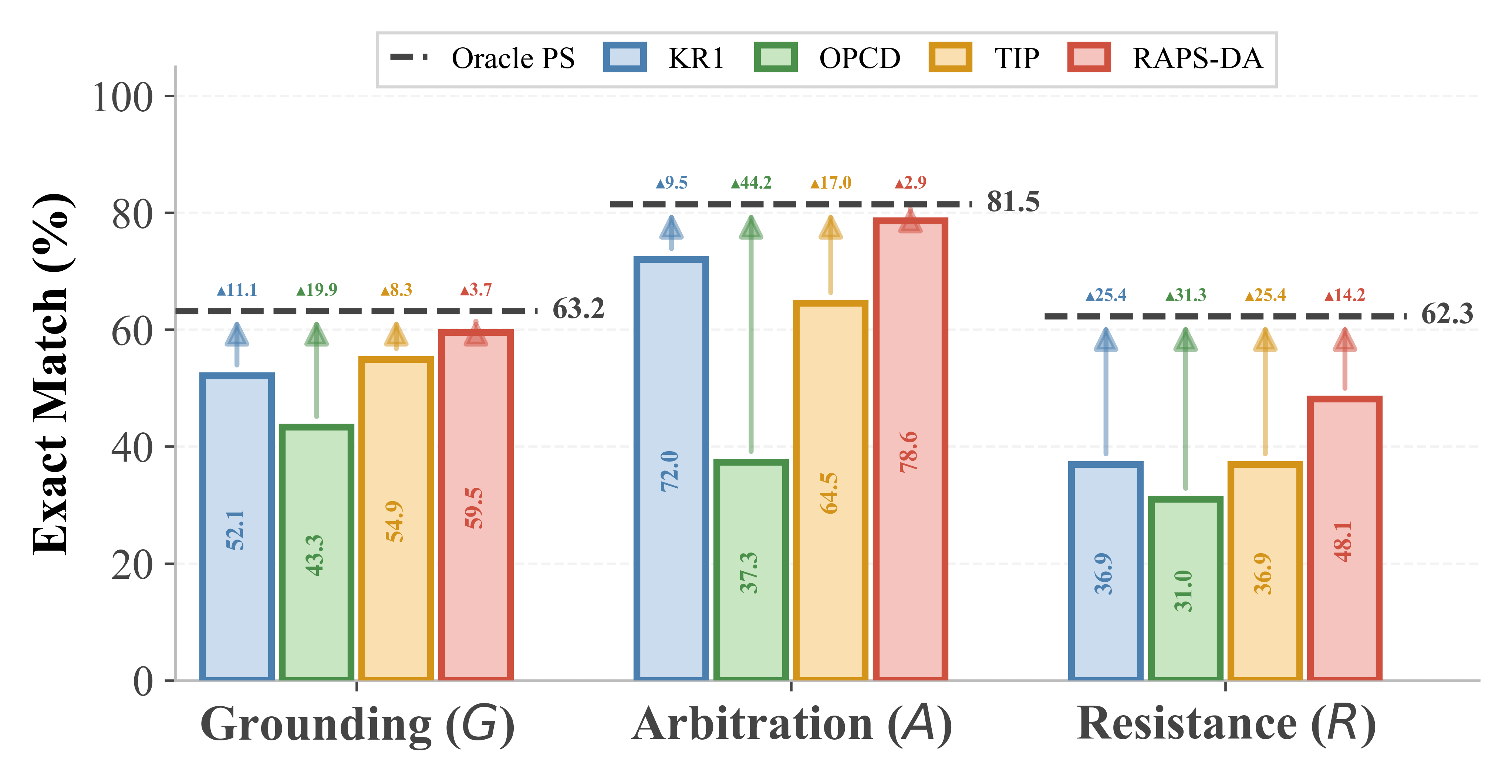}
\caption{Regime-level performance comparison (Qwen-7B). Bars show the regime-averaged EM of four representative methods; the dashed line marks the oracle peer specialist upper bound. $\blacktriangle$ values indicate the remaining gap to the oracle. \ours{} achieves the smallest gap across all three regimes.}
\label{fig:main-regime}
\end{figure}

\paragraph{\ours{} outperforms all baseline categories without trading off across regimes.}
Prompting and decoding methods (Astute-RAG, CK-PLUG) lack training signal for conflict resolution and fail on adversarial context, with Resistance averages below 11\% on Qwen.
SFT and RL methods (Mixed SFT, GRPO, KR1) improve Resistance substantially through conflict-aware training objectives but share a single policy for all regimes, resulting in cross-regime interference that limits their Grounding average to around 50\%.
OPD methods (OPCD, TIP) learn from teacher demonstrations and achieve competitive Grounding performance, but a single generalist teacher cannot specialize to all three regimes simultaneously, limiting their Resistance average.
\ours{} addresses these limitations by combining regime-specialized peer teachers with token-level selection.
On the Qwen backbone, \ours{} achieves regime averages of 59.5\% ($G$), 78.6\% ($A$), and 48.1\% ($R$), improving over KR1 by 7.4, 6.6, and 11.2 points and over TIP by 4.6, 14.0, and 11.2 points, respectively.
The largest margin in both comparisons appears on Resistance, where the model must reject adversarial context entirely, confirming that regime-specialized teachers provide the greatest benefit when the required behavior diverges most from standard RAG.
Crucially, unlike baselines that trade Grounding accuracy for Resistance robustness, \ours{} improves both ends of the reliability spectrum simultaneously. On S1 (correct context), accuracy reaches 89.3\% and 87.0\% on PC-MR and PC-MC, while on S2 (adversarial context) it improves to 51.5\%, 48.2\%, and 44.6\%, demonstrating that learning to reject misleading context does not reduce the ability to leverage correct context.

\paragraph{Regime-specialized OPD narrows the gap to the oracle upper bound.}
Figure~\ref{fig:main-regime} visualizes the regime-level comparison against the oracle peer specialist, a non-deployable upper bound that requires ground-truth regime labels at inference.
\ours{} closes 69.6\% of the gap between KR1 and the oracle on Arbitration, and even surpasses the oracle on Musique within the Grounding regime, suggesting that cross-regime knowledge transfer can compensate for the oracle's lack of inter-regime information sharing.
Among OPD baselines, TIP achieves competitive Grounding through importance-weighted token selection but remains limited on Resistance due to its single generalist teacher. OPCD, which also serves as our single-teacher OPD ablation (Table~\ref{tab:ablation-main}), underperforms substantially on Arbitration and Resistance, confirming that a jointly-trained teacher systematically underfits every regime.
These results indicate that the gains originate from regime-specialized supervision quality rather than from model scale or ensemble effects.

\subsection{Out-of-Distribution Generalization}\label{sec:ood}

We next evaluate whether the conflict-resolution strategy learned by \ours{} transfers to held-out benchmarks not seen during teacher training or student training.
ConflictQA tests robustness under parametric--contextual conflict; TruthfulQA tests parametric-knowledge preservation in a no-context setting, verifying that training does not erode the base model's truthfulness.
Table~\ref{tab:ood} reports results on both benchmarks.

\begin{table}[t]
\centering
\small
\caption{Out-of-distribution evaluation on ConflictQA (EM, \%) and TruthfulQA (EM, \%, no-context). Neither benchmark participates in any training stage. Best in \textbf{bold}, second best \underline{underlined}.}
\label{tab:ood}
\begin{tabular}{lcc}
\toprule
Method & ConflictQA & TruthfulQA \\
\midrule
RAG prompting                              & 34.3  & 4.4 \\
Astute-RAG~\cite{wang2024astute}           & 16.9 & \textbf{5.4} \\
CK-PLUG~\cite{bi2025parametersvscontextfinegrained} & 7.9  & 0.0 \\
Mixed SFT                                  & 39.7 & 1.8 \\
GRPO w/ RAG                                & 36.3 & 4.2 \\
Knowledgeable-R1~\cite{lin2025knowledgeable} & 36.5 & 4.2 \\
OPCD~\cite{ye2026opcd}                     & \underline{40.0} & 3.7 \\
TIP~\cite{tip2026}                         & 38.4 & 4.2 \\
\textbf{\ours{} (ours)}                   & \textbf{42.7} & \underline{4.4} \\
\bottomrule
\end{tabular}
\end{table}

\ours{} achieves the highest ConflictQA EM of 42.7\%, outperforming all baselines including OPCD (40.0\%) and TIP (38.4\%), while maintaining 4.4\% on TruthfulQA, matching the base model and confirming that regime-specialized training does not erode parametric knowledge.
Notably, OPCD attains comparable ConflictQA performance but drops to 3.7\% on TruthfulQA, suggesting that its context-conditioned teacher transfers conflict-handling ability at the cost of parametric knowledge retention.
Two edge cases merit discussion. CK-PLUG degenerates to 0.0\% on TruthfulQA because the no-context setting leaves its confidence-gain signal undefined. Astute-RAG achieves the highest TruthfulQA score of 5.4\% but the lowest ConflictQA score among training-based methods at 16.9\%, as its prompting-based consolidation lacks the learned conflict resolution needed for parametric--contextual conflicts.

\subsection{Component and Motivation-Validating Ablations}\label{sec:ablation}

The ablations in this section serve two complementary purposes.
First, they verify that each component of \ours{} is necessary by removing it and measuring the resulting performance drop.
Second, they validate the underlying motivations that justify the framework's design, including the presence of cross-regime interference under a single teacher, the complementary roles of sample-level routing and token-level selection, the necessity of difficulty annealing over static masking, and the individual contributions of the three diagnostic signals.
Table~\ref{tab:ablation-main} summarizes all motivation-validating variants and component removals; subsequent subsections analyze each finding in detail.

\begin{table}[t]
\centering
\setlength{\tabcolsep}{2.8pt}
\renewcommand{\arraystretch}{1.12}
\caption{Motivation-validating ablations on Qwen2.5-7B-Instruct. Each row removes or replaces a single design choice while keeping all other components and hyperparameters fixed. The upper block validates our core design motivations; the lower block isolates individual selector components. Best results are in \textbf{bold}; \colorbox{teal!12}{shaded row} denotes the full method.}
\label{tab:ablation-main}
\footnotesize
\resizebox{\linewidth}{!}{%
\begin{tabular}{l ccc ccc c c ccc | ccc}
\toprule
& \multicolumn{3}{c}{\textbf{S1 Correct ($G$)}} & \multicolumn{3}{c}{\textbf{S2 Wrong ($R$)}} & \textbf{S3 ($A$)} & \textbf{S4 ($A$)} & \multicolumn{3}{c}{\textbf{S5 Partly-Irr. ($G$)}} & \multicolumn{3}{c}{\textbf{Regime Avg.}}\\
\cmidrule(lr){2-4}\cmidrule(lr){5-7}\cmidrule(lr){8-8}\cmidrule(lr){9-9}\cmidrule(lr){10-12}\cmidrule(lr){13-15}
\textbf{Variant} & PC-MR & PC-MC & PC-QA & NC-MR & NC-MC & NC-QA & SC & ExpPE & HotPot & 2Wiki & Musique & $G$ & $A$ & $R$\\
\midrule
\rowcolor{teal!12}
\textbf{\ours{} (full)} & \textbf{89.3} & \textbf{87.0} & \textbf{80.9} & \textbf{51.5} & \textbf{48.2} & \textbf{44.6} & \textbf{87.3} & \textbf{69.9} & \textbf{33.9} & \textbf{45.5} & \textbf{20.1} & \textbf{59.5} & \textbf{78.6} & \textbf{48.1} \\
\midrule
\multicolumn{15}{l}{\emph{Motivation-validating variants}} \\
Single-teacher OPD (= OPCD)                 & 64.7 & 64.4 & 62.8 & 40.2 & 37.0 & 15.8 & 56.5 & 18.1 & 24.6 & 34.5 & 8.8 & 43.3 & 37.3 & 31.0 \\
Multi-teacher uniform pooling      & 80.7 & 78.4 & 84.7 & 15.3 & 10.4 & 19.2 & 67.4 & 55.3 & 33.5 & 44.4 & 11.1 & 55.5 & 61.4 & 15.0 \\
Routing only                       & 83.3 & 80.7 & 85.2 & 24.2 & 19.1 & 34.9 & 75.1 & 65.9 & 33.5 & 45.9 & 13.8 & 57.1 & 70.5 & 26.1 \\
Token selection only               & 77.4 & 75.5 & 79.0 & 21.0 & 14.1 & 25.8 & 65.2 & 64.1 & 30.9 & 35.0 & 13.1 & 51.8 & 64.7 & 20.3 \\
Static masking                     & 85.5 & 84.3 & 83.0 & 40.7 & 40.5 & 41.2 & 82.2 & 69.9 & 32.9 & 43.9 & 15.1 & 57.4 & 76.1 & 40.8 \\
\midrule
\multicolumn{15}{l}{\emph{Component removals}} \\
$-$ Soft weight $\omega$            & 88.3 & 87.5 & 84.3 & 41.8 & 39.0 & 44.6 & 87.2 & 69.8 & 33.0 & 44.8 & 19.4 & 59.6 & 78.5 & 41.8 \\
$-$ Information filter               & 89.1 & 86.8 & 84.2 & 45.6 & 41.0 & 42.0 & 86.1 & 72.4 & 33.6 & 45.5 & 18.2 & 59.6 & 79.2 & 42.9 \\
$-$ Stability filter                 & 88.4 & 87.7 & 81.1 & 46.8 & 40.0 & 43.6 & 87.1 & 69.1 & 32.8 & 43.6 & 18.1 & 58.6 & 78.1 & 43.5 \\
\bottomrule
\end{tabular}}
\end{table}

Several observations emerge from Table~\ref{tab:ablation-main}.
\emph{Single-teacher OPD}, equivalent to OPCD in Table~\ref{tab:main-results}, removes both regime routing and token selection. It suffers the largest overall degradation, with Arbitration dropping from 78.6\% to 37.3\% and Resistance from 48.1\% to 31.0\%. This confirms that a single jointly-trained teacher cannot provide effective supervision across heterogeneous conflict regimes.
\emph{Multi-teacher uniform pooling} recovers Grounding performance to 55.5\% by averaging multiple teacher signals, but catastrophically collapses on Resistance to 15.0\%, indicating that uniformly mixing teacher distributions re-introduces cross-regime interference at the token level.
Comparing \emph{Routing only} and \emph{Token selection only} reveals that the two mechanisms are complementary rather than substitutable. Routing alone improves Arbitration to 70.5\% but leaves Resistance at 26.1\%, while token selection alone achieves 64.7\% on Arbitration but only 20.3\% on Resistance. Neither mechanism in isolation approaches the full model.
\emph{Static masking}, which retains a fixed set of tokens without annealing, achieves 40.8\% on Resistance compared to 48.1\% for the full model. The 7.3-point gap confirms that the progressive easy-to-hard schedule contributes meaningfully beyond a fixed selection budget.
Among the component removals, the soft weight $\omega$ has the largest impact on Resistance, reducing it by 6.3 points. This is consistent with its role in prioritizing confidently misaligned tokens, which are most prevalent in the adversarial regime.

\subsubsection{Cross-Regime Interference and the Need for Multiple Teachers}\label{sec:abl-interference}

A central design motivation for \ours{} is that a single model struggles to simultaneously integrate reliable context, arbitrate mixed evidence, and reject adversarial passages, because these behaviors impose partially contradictory optimization pressures.
To make this interference empirically visible, we train three regime-specialized peer teachers in isolation and evaluate each teacher on all three regimes.
Table~\ref{tab:cross-regime} reports the resulting cross-regime evaluation matrix.

\begin{table}[t]
\centering
\small
\caption{Cross-regime evaluation matrix. Each row corresponds to a teacher trained exclusively on one regime; columns indicate the evaluation regime. Diagonal entries (bold) reflect specialization, while off-diagonal drops quantify cross-regime interference.}
\label{tab:cross-regime}
\begin{tabular}{l|ccc|c}
\toprule
Trained on $\backslash$ Eval on & $G$ & $A$ & $R$ & Avg \\
\midrule
Specialist on $G$  & \textbf{60.5} & 59.7 & 11.0 & 43.7 \\
Specialist on $A$  & 57.7 & \textbf{77.7} & 30.5 & 55.3 \\
Specialist on $R$  & 35.8 & 53.2 & \textbf{47.2} & 45.4 \\
\bottomrule
\end{tabular}
\end{table}
%% 原始逐场景数据备注：
%%   run_v5e_regime_s2:  52.5 | 52.0 | 28.6 | 51.7 | 42.8 | 47.0 | 65.0 | 41.4 | 29.0 | 38.6 | 14.3
%%   v5f_regime_s1s5:    88.1 | 88.3 | 88.6 | 14.3 | 11.2 |  7.4 | 72.0 | 47.4 | 33.7 | 46.2 | 18.4
%%   v5g_regime_s3s4:    82.7 | 81.8 | 87.8 | 29.0 | 29.3 | 33.1 | 83.3 | 72.1 | 32.7 | 45.1 | 15.8
%%   (列顺序: PC-MR | PC-MC | PC-QA | NC-MR | NC-MC | NC-QA | SC | ExpPE | HotPot | 2Wiki | Musique)

The matrix exhibits a pronounced diagonal structure.
The Grounding specialist achieves 60.5\% on its own regime but collapses to 11.0\% on Resistance, confirming that optimizing for context integration directly conflicts with context rejection.
Conversely, the Resistance specialist attains 47.2\% on adversarial samples but drops to 35.8\% on Grounding, where its rejection bias suppresses useful evidence.
The Arbitration specialist reaches the highest single-regime score of 77.7\% but still drops substantially on Resistance (30.5\%), further illustrating that no single teacher can cover all regimes.
These results validate the core premise that a single supervisory policy cannot simultaneously serve all conflict regimes, and that regime-specific teachers are necessary to eliminate cross-regime interference.

\subsubsection{Why Difficulty Annealing Matters}\label{sec:abl-annealing}

\begin{figure}[t]
\centering
\includegraphics[width=\linewidth]{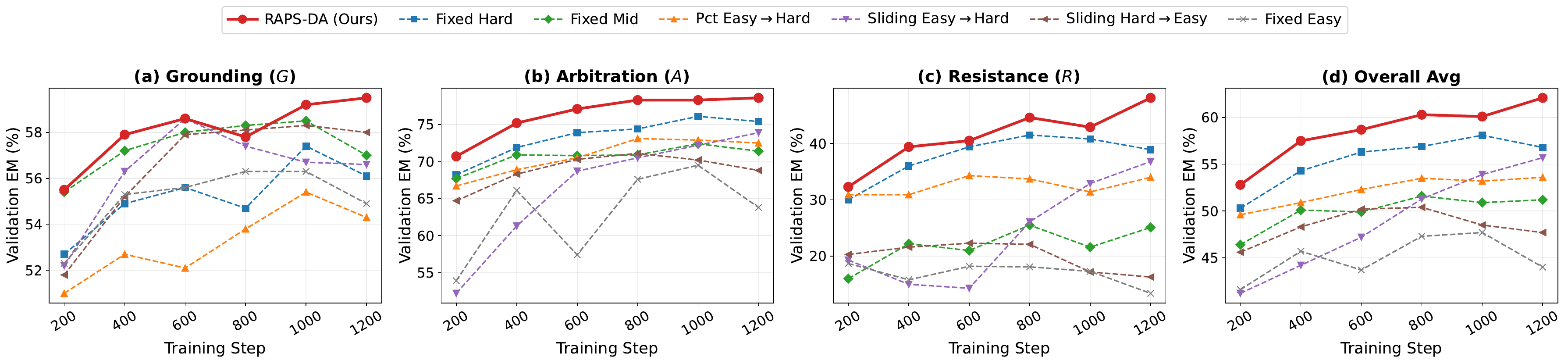}
\caption{Training dynamics under different annealing schedules. Each panel reports validation EM over training steps for one regime or the overall average. \ours{} with linear annealing (solid orange) reaches the highest final plateau across all regimes. On Resistance, Sliding Hard$\to$Easy collapses to 16.3\% and Fixed Easy to 13.4\%, confirming that the easy-to-hard ordering is essential for learning adversarial context rejection.}
\label{fig:annealing-dynamics}
\end{figure}

The difficulty annealing in \ours{} progressively narrows the retained token set from broad distributional coverage to the most difficult tokens.
To isolate the contribution of the annealing schedule from the token budget, we compare against two families of alternatives that control for the same 50\% retention ratio.
The first family, \emph{static masking}, retains a fixed set of tokens at a constant difficulty level throughout training, with three variants targeting hard, medium, and easy tokens respectively.
The second family, \emph{sliding-window} schedules, shifts a fixed-width 50\% window across the difficulty spectrum over training, in either the easy-to-hard or the hard-to-easy direction.
We additionally include a curriculum-based variant that gradually expands the retained difficulty percentile range.
If the performance gain stems purely from the reduced token budget, static variants should match \ours{}; if the ordering of token exposure matters, the easy-to-hard direction should outperform the reverse.
Figure~\ref{fig:annealing-dynamics} visualizes the validation EM trajectories across all variants over training.

%% 绘图脚本：figs/plot_annealing_dynamics.py
%% 输出文件：figs/annealing_dynamics.pdf
%% 数据来源：Per-regime EM, step 200-1200, 每 200 step 一个点
%%   ours (G):            55.5 | 57.9 | 58.6 | 57.8 | 59.2 | 59.5
%%   ours (A):            70.7 | 75.2 | 77.1 | 78.3 | 78.3 | 78.6
%%   ours (R):            32.3 | 39.4 | 40.5 | 44.6 | 42.9 | 48.1
%%   ours (AVG):          52.8 | 57.5 | 58.7 | 60.3 | 60.1 | 62.1
%%   v5a (Pct E→H, AVG):  49.6 | 50.9 | 52.3 | 53.5 | 53.2 | 53.6
%%   v5c (Slide E→H, AVG):41.2 | 44.2 | 47.2 | 51.3 | 53.9 | 55.7
%%   v5d (Slide H→E, AVG):45.6 | 48.3 | 50.2 | 50.4 | 48.5 | 47.7
%%   v5h (Fixed Hard, AVG):50.3 | 54.3 | 56.3 | 56.9 | 58.1 | 56.8
%%   v5i (Fixed Easy, AVG):41.6 | 45.7 | 43.7 | 47.3 | 47.7 | 44.0
%%   v5j (Fixed Mid, AVG): 46.4 | 50.1 | 49.9 | 51.6 | 50.9 | 51.2

\begin{table}[t]
\centering
\setlength{\tabcolsep}{2.6pt}
\renewcommand{\arraystretch}{1.12}
\caption{Model-scale generalization on Qwen2.5-Instruct at 3B and 14B. EM (\%) is reported for all five scenarios and three regime averages. KR1 baseline numbers are reproduced from \citet{lin2025knowledgeable}; ``---'' denotes results not reported in the original paper. The 7B results appear in Table~\ref{tab:main-results}. \colorbox{teal!12}{Shaded rows} denote our method.}
\label{tab:scaling}
\footnotesize
\resizebox{\linewidth}{!}{%
\begin{tabular}{l ccc ccc c c ccc | ccc}
\toprule
& \multicolumn{3}{c}{\textbf{S1 Correct ($G$)}} & \multicolumn{3}{c}{\textbf{S2 Wrong ($R$)}} & \textbf{S3 ($A$)} & \textbf{S4 ($A$)} & \multicolumn{3}{c}{\textbf{S5 Partly-Irr. ($G$)}} & \multicolumn{3}{c}{\textbf{Regime Avg.}}\\
\cmidrule(lr){2-4}\cmidrule(lr){5-7}\cmidrule(lr){8-8}\cmidrule(lr){9-9}\cmidrule(lr){10-12}\cmidrule(lr){13-15}
\textbf{Method} & PC-MR & PC-MC & PC-QA & NC-MR & NC-MC & NC-QA & SC & ExpPE & HotPot & 2Wiki & Musique & $G$ & $A$ & $R$\\
\midrule
\multicolumn{15}{l}{\texttt{Qwen2.5-3B-Instruct}} \\
Query-only        & 16.8 & 18.4 & 23.9 & 16.0 & 18.3 & 22.6 & 20.8 & 53.3 & 12.6 & 11.3 & 2.1  & 14.2 & 37.0 & 19.0 \\
RAG prompting     & 55.1 & 52.9 & 59.5 & 13.5 & 8.7  & 6.8  & 51.2 & 42.2 & 14.7 & 13.0 & 3.6  & 33.1 & 46.7 & 9.7 \\
GRPO w/ RAG       & 71.0 & 70.6 & 80.0 & 19.9 & 15.1 & 19.4 & 66.7 & 53.6 & 24.8 & 35.6 & 9.1  & 48.5 & 60.1 & 18.1 \\
KR1               & 66.2 & 59.8 & 78.7 & 28.8 & 28.6 & 22.0 & ---   & 54.7 & ---   & ---   & ---   & ---   & ---   & 26.5 \\
\rowcolor{teal!12}
\textbf{\ours{} (ours)} & 74.8 & 73.5 & 83.2 & 35.6 & 33.3 & 28.7 & 69.8 & 57.4 & 28.4 & 39.9 & 12.1 & 52.0 & 63.6 & 32.5 \\
\midrule
\multicolumn{15}{l}{\texttt{Qwen2.5-14B-Instruct}} \\
Query-only        & 30.0 & 26.5 & 35.8 & 30.6 & 28.9 & 39.4 & 30.9 & 70.8 & 25.4 & 26.7 & 5.7  & 25.0 & 50.9 & 33.0 \\
RAG prompting     & 63.2 & 63.9 & 73.3 & 22.2 & 13.0 & 28.1 & 61.9 & 70.5 & 22.9 & 22.5 & 6.9  & 42.1 & 66.2 & 21.1 \\
GRPO w/ RAG       & 73.6 & 74.7 & 78.9 & 38.9 & 31.7 & 36.5 & 74.0 & ---   & 34.7 & 39.0 & 14.7 & ---   & ---   & 35.7 \\
KR1               & 70.6 & 75.2 & 78.8 & 47.8 & 33.1 & 38.1 & 72.5 & ---   & 37.0 & 42.3 & 14.6 & ---   & ---   & 39.7 \\
\rowcolor{teal!12}
\textbf{\ours{} (ours)} & 84.4 & 85.7 & 88.2 & 60.3 & 55.8 & 42.6 & 78.5 & 75.3 & 38.5 & 47.8 & 17.6 & 60.4 & 76.9 & 52.9 \\
\bottomrule
\end{tabular}}
\end{table}

Two consistent findings emerge from Figure~\ref{fig:annealing-dynamics}.
First, the ordering of token exposure matters substantially more than the token budget.
All variants retain approximately the same fraction of tokens, yet their final overall EM spans an 18.1-point range (62.1 vs.\ 44.0).
The Sliding Hard$\to$Easy schedule exposes the student to the most difficult tokens first and then gradually relaxes to easier ones. It peaks at 50.4\% overall at step 800 then declines to 47.7\%, well below Fixed Hard's peak of 58.1\%.
This confirms that the student requires broad distributional coverage in early training to establish regime-consistent foundations before supervision can safely narrow to high-conflict positions.
Second, the Resistance regime exhibits the greatest sensitivity to annealing design.
Fixed Easy tokens yield a final EM of only 13.4\% on Resistance, Fixed Hard tokens reach 38.9\%, and the linear annealing schedule of \ours{} achieves 48.1\%.
The linear schedule reaches the highest final plateau on all four panels, whereas the Sliding Hard$\to$Easy trajectory on Resistance begins declining after step 600, dropping from 22.3\% to 16.3\%, indicating that premature hard-token exposure leads to unstable optimization.

\subsection{Model-Scale Generalization}\label{sec:scaling}

To verify that the benefits of regime-aware peer specialization are not restricted to a single model scale, we evaluate \ours{} on Qwen2.5-Instruct at 3B and 14B in addition to the 7B results reported in Table~\ref{tab:main-results}.
Table~\ref{tab:scaling} reports per-scenario EM alongside the KR1 and GRPO baselines, with KR1 numbers reproduced from \citet{lin2025knowledgeable}.

\ours{} yields consistent improvements at both model scales.
On the 3B backbone, \ours{} achieves regime averages of 52.0\%, 63.6\%, and 32.5\% on Grounding, Arbitration, and Resistance, respectively. Compared with KR1, the gains on Resistance are 6.0 points, and on Grounding the per-scenario improvements range from 4.5 to 8.6 points across PC-MR, PC-MC, and PC-QA.
This is consistent with the expectation that smaller models, having weaker parametric knowledge, benefit more from regime-specialized supervision.
On the 14B backbone, \ours{} improves over KR1 by 13.2 points on Resistance and achieves 76.9\% on Arbitration, demonstrating that the approach remains effective even when the base model already possesses stronger knowledge.
Notably, the Resistance regime shows the largest relative gains at both scales, reinforcing the finding from Section~\ref{sec:ablation} that adversarial context rejection benefits most from regime-specialized training signals.

\section{Conclusion}\label{sec:conclusion}

We presented \ours{}, a regime-aware peer specialization framework for robust retrieval-augmented generation under heterogeneous knowledge conflicts.
Rather than treating knowledge conflict as a single phenomenon, \ours{} decomposes it into distinct reliability regimes and provides regime-matched supervision through same-scale peer specialists. Combined with conflict-aware token selection and difficulty annealing, this enables more targeted and stable learning under conflicting evidence.
Experiments across multiple conflict benchmarks show that \ours{} consistently outperforms prompting, decoding, fine-tuning, RL, and single-teacher OPD baselines.
The resulting student generalizes across conflict scenarios, model scales, and architectures, while approaching the performance of an oracle regime-specialist upper bound. 
Ablation studies further demonstrate that the three components are complementary and jointly responsible for the observed gains.
More broadly, our results suggest that robust conflict resolution is fundamentally a specialization problem rather than a scaling problem. 
When different reliability regimes require different behaviors, training signals should reflect this heterogeneity instead of forcing a single supervision policy to reconcile incompatible objectives. 
We hope this perspective motivates future work on regime-aware training for retrieval-augmented language models.

\bibliographystyle{plainnat}
\bibliography{references}

@String(NIPS= {Adv. Neural Inform. Process. Syst.})

@String(ICLR = {Int. Conf. Learn. Represent.})

@String(NIPS  = {NeurIPS})

@String(ICLR  = {ICLR})

@string(NIPS= {Adv. Neural Inform. Process. Syst.})

@string(ICLR = {Int. Conf. Learn. Represent.})

@string(ICML = {Int. Conf. Mach. Learn.})

@string(COLM = {Conference on Language Modeling})

@string(EMNLP = {Conf. Empir. Methods Nat. Lang. Process.})

@string(ACL ={Proc. Annu. Meet. Assoc. Comput Linguist.})

@string(NAACL ={Proc. Conf. North American Chapter Assoc. Comput. Linguist.})

@string(COLING ={Int. Conf. Comput. Linguist., Lang. Resour. Eval.})

@article{guo2025deepseekr1_nature,
  title   = {DeepSeek-R1 incentivizes reasoning in {LLMs} through reinforcement learning},
  author  = {Guo, Daya and Yang, Dejian and Zhang, Haowei and Song, Junxiao and Wang, Peiyi and Zhu, Qihao and Xu, Runxin and Zhang, Ruoyu and Ma, Shirong and Bi, Xiao and others},
  journal = {Nature},
  year    = {2025},
  volume  = {645},
  number  = {8081},
  pages   = {633-638},
  month   = {sep},
  
}

@article{shao2024deepseekmath,
  title={Deepseekmath: Pushing the limits of mathematical reasoning in open language models},
  author={Shao, Zhihong and Wang, Peiyi and Zhu, Qihao and Xu, Runxin and Song, Junxiao and Bi, Xiao and Zhang, Haowei and Zhang, Mingchuan and Li, YK and others},
  journal={arXiv preprint arXiv:2402.03300},
  year={2024}
}

@inproceedings{he2024retrieving,
  title={Retrieving, Rethinking and Revising: The Chain-of-Verification Can Improve Retrieval Augmented Generation},
  author={He, Bolei and Chen, Nuo and He, Xinran and Yan, Lingyong and Wei, Zhenkai and Luo, Jinchang and Ling, Zhen-Hua},
  booktitle=EMNLP,
  pages={10371--10393},
  year={2024}
}

@inproceedings{xu2024knowledge,
  title={Knowledge Conflicts for LLMs: A Survey},
  author={Xu, Rongwu and Qi, Zehan and Guo, Zhijiang and Wang, Cunxiang and Wang, Hongru and Zhang, Yue and Xu, Wei},
  booktitle=EMNLP,
  pages={8541--8565},
  year={2024}
}

@inproceedings{jin2024tug,
  title={Tug-of-War between Knowledge: Exploring and Resolving Knowledge Conflicts in Retrieval-Augmented Language Models},
  author={Jin, Zhuoran and Cao, Pengfei and Chen, Yubo and Liu, Kang and Jiang, Xiaojian and Xu, Jiexin and Qiuxia, Li and Zhao, Jun},
  booktitle=COLING,
  pages={16867--16878},
  year={2024}
}

@inproceedings{su2024conflictbank,
  title={CONFLICTBANK: a benchmark for evaluating knowledge conflicts in large language models},
  author={Su, Zhaochen and Zhang, Jun and Qu, Xiaoye and Zhu, Tong and Li, Yanshu and Sun, Jiashuo and Li, Juntao and Zhang, Min and Cheng, Yu},
  booktitle=NIPS,
  pages={103242--103268},
  year={2024}
}

@inproceedings{Xie2024KnowledgeConflict,
  title={Adaptive Chameleon or Stubborn Sloth: Revealing the Behavior of Large Language Models in Knowledge Conflicts},
  author={Xie, Jian and Zhang, Kai and Chen, Jiangjie and Lou, Renze and Su, Yu},
  booktitle=ICLR,
  year={2024},
}

@article{wang2023resolving,
  title={Resolving Knowledge Conflicts in Large Language Models},
  author={Wang, Yike and Feng, Shangbin and Wang, Heng and Shi, Weijia and Balachandran, Vidhisha and He, Tianxing and Tsvetkov, Yulia},
  year={2024},
  journal=COLM
}

@article{wang2024astute,
  title={Astute rag: Overcoming imperfect retrieval augmentation and knowledge conflicts for large language models},
  author={Wang, Fei and Wan, Xingchen and Sun, Ruoxi and Chen, Jiefeng and Ar{\i}k, Sercan {\"O}},
  journal=ACL,
  year={2025}
}

@inproceedings{fang2024raat,
  title        = {Enhancing Noise Robustness of Retrieval-Augmented Language Models via {RAAT}},
  author       = {Fang, Yucheng and Wang, Ruochen and Qian, Kun and Feng, Yansong and Yang, Diyi and He, He},
  booktitle    = ACL,
  year         = {2024},
}

@inproceedings{wang2025ramdocs,
  title        = {Retrieval-Augmented Generation with Conflicting Evidence},
  author       = {Wang, Han and Prasad, Archiki and Stengel-Eskin, Elias and Bansal, Mohit},
  booktitle    = COLM,
  year         = {2025},  
}

@inproceedings{asai2024self,
  title={Self-RAG: Learning to Retrieve, Generate, and Critique through Self-Reflection},
  author={Asai, Akari and Wu, Zeqiu and Wang, Yizhong and Sil, Avi and Hajishirzi, Hannaneh},
  booktitle=ICLR,
  year={2024}
}

@article{ding2024retrieve,
  title={Retrieve only when it needs: Adaptive retrieval augmentation for hallucination mitigation in large language models},
  author={Ding, Hanxing and Pang, Liang and Wei, Zihao and Shen, Huawei and Cheng, Xueqi},
  journal={arXiv preprint arXiv:2402.10612},
  year={2024}
}

@misc{bi2025parametersvscontextfinegrained,
      title={Parameters vs. Context: Fine-Grained Control of Knowledge Reliance in Language Models}, 
      author={Baolong Bi and Shenghua Liu and Yiwei Wang and Yilong Xu and Junfeng Fang and Lingrui Mei and Xueqi Cheng},
      year={2025},
      eprint={2503.15888},
      archivePrefix={arXiv},
      primaryClass={cs.CL}, 
}

@article{jin2025search,
  title={Search-r1: Training llms to reason and leverage search engines with reinforcement learning},
  author={Jin, Bowen and Zeng, Hansi and Yue, Zhenrui and Yoon, Jinsung and Arik, Sercan and Wang, Dong and Zamani, Hamed and Han, Jiawei},
  journal={arXiv preprint arXiv:2503.09516},
  year={2025}
}

@misc{qwen2025qwen25technicalreport,
      title={Qwen2.5 Technical Report}, 
      author={Qwen and : and An Yang and Baosong Yang and Beichen Zhang and Binyuan Hui and Bo Zheng and Bowen Yu and Chengyuan Li and Dayiheng Liu and Fei Huang and Haoran Wei and Huan Lin and Jian Yang and Jianhong Tu and Jianwei Zhang and Jianxin Yang and Jiaxi Yang and Jingren Zhou and Junyang Lin and Kai Dang and Keming Lu and Keqin Bao and Kexin Yang and Le Yu and Mei Li and Mingfeng Xue and Pei Zhang and Qin Zhu and Rui Men and Runji Lin and Tianhao Li and Tianyi Tang and Tingyu Xia and Xingzhang Ren and Xuancheng Ren and Yang Fan and Yang Su and Yichang Zhang and Yu Wan and Yuqiong Liu and Zeyu Cui and Zhenru Zhang and Zihan Qiu},
      year={2025},
      eprint={2412.15115},
      archivePrefix={arXiv},
      primaryClass={cs.CL},
     
}

@inproceedings{shi2024cad,
  title={Trusting Your Evidence: Hallucinate Less with Context-Aware Decoding},
  author={Shi, Weijia and Han, Xiaochuang and Lewis, Mike and Tsvetkov, Yulia and Zettlemoyer, Luke and Yih, Scott Wen-tau},
  booktitle=NAACL,
  pages={783--800},
  year={2024}
}

@inproceedings{chuang2024dola,
  title={DoLa: Decoding by Contrasting Layers Improves Factuality in Large Language Models},
  author={Chuang, Yung-Sung and Xie, Yujia and Luo, Hongyin and Kim, Yoon and Glass, James and He, Pengcheng},
  booktitle=ICLR,
  year={2024}
}

@article{zhang2024dccd,
  title={Dynamic Contrastive Decoding for Knowledge Conflict Resolution in Large Language Models},
  author={Zhang, Xueying and Chen, Yanqiu and Li, Yongkang},
  journal={arXiv preprint arXiv:2405.13183},
  year={2024}
}

@article{xu2024inforag,
  title={InFO-RAG: Information-Filtered On-Policy Retrieval-Augmented Generation},
  author={Xu, Chenliang and Guo, Jiaxin and Wang, Yiwei and Liu, Shenghua},
  journal={arXiv preprint arXiv:2406.19009},
  year={2024}
}

@article{li2024kaft,
  title={Knowledge-Aware Fine-Tuning for Robust Retrieval-Augmented Generation},
  author={Li, Xiaoyu and Zhang, Hao and Wang, Zhiyuan},
  journal={arXiv preprint arXiv:2407.12854},
  year={2024}
}

@article{lin2025knowledgeable,
  title={Knowledgeable-R1: Reinforcement Learning for Knowledge-Conflict Resolution in RAG},
  author={Lin, Zhen and Wang, Yifei and Chen, Hao and Liu, Zhiyuan},
  journal={arXiv preprint arXiv:2503.12345},
  year={2025}
}

@inproceedings{agarwal2024gkd,
  title={On-Policy Distillation of Language Models: Learning from Self-Generated Mistakes},
  author={Agarwal, Rishabh and Vieillard, Nino and Stanczyk, Piotr and Ramos, Sabela and Geist, Matthieu and Bachem, Olivier},
  booktitle=ICLR,
  year={2024}
}

@inproceedings{gu2024minillm,
  title={MiniLLM: Knowledge Distillation of Large Language Models},
  author={Gu, Yuxian and Dong, Li and Wei, Furu and Huang, Minlie},
  booktitle=ICLR,
  year={2024}
}

@article{fu2026revisiting,
  title={Revisiting On-Policy Distillation: Three Failure Modes and Top-K Truncated Reverse-KL},
  author={Fu, Yao and others},
  journal={arXiv preprint arXiv:2603.25562},
  year={2026}
}

@article{luo2026stable,
  title={Stable On-Policy Distillation: Mitigating Length Inflation in LLM Training},
  author={Luo, Haoran and others},
  journal={arXiv preprint arXiv:2604.08527},
  year={2026}
}

@article{tip2026,
  title={TIP: Token Importance Profiling for Efficient On-Policy Distillation},
  author={Anonymous},
  journal={arXiv preprint arXiv:2604.14084},
  year={2026}
}

@article{scope2026,
  title={SCOPE: Correctness-Based Dual-Path Token Weighting for On-Policy Distillation},
  author={Anonymous},
  journal={arXiv preprint arXiv:2604.10688},
  year={2026}
}

@article{srpo2026,
  title={SRPO: Self-Refined Policy Optimization via Correctness-Aware Routing},
  author={Anonymous},
  journal={arXiv preprint arXiv:2604.02288},
  year={2026}
}

@article{yang2026gopd,
  title={G-OPD: Generalized On-Policy Distillation as Dense KL-Constrained Reinforcement Learning},
  author={Yang, Zichun and others},
  journal={arXiv preprint arXiv:2602.12125},
  year={2026}
}

@article{wu2026lightning,
  title={Lightning On-Policy Distillation: Teacher Consistency is All You Need},
  author={Wu, Chengyue and others},
  journal={arXiv preprint arXiv:2604.13010},
  year={2026}
}

@article{ye2026opcd,
  title={On-Policy Context Distillation for Language Models},
  author={Ye, Tianzhu and Dong, Li and Wu, Xun and Huang, Shaohan and Wei, Furu},
  journal={arXiv preprint arXiv:2602.12275},
  year={2026}
}

@article{zhao2026opsd,
  title={Privileged Information Distillation for Language Models},
  author={Penaloza, Emiliano and Vattikonda, Dheeraj and Gontier, Nicolas and Lacoste, Alexandre and Charlin, Laurent and Caccia, Massimo},
  journal={arXiv preprint arXiv:2602.04942},
  year={2026}
}

@article{hinton2015distilling,
  title={Distilling the Knowledge in a Neural Network},
  author={Hinton, Geoffrey and Vinyals, Oriol and Dean, Jeff},
  journal={arXiv preprint arXiv:1503.02531},
  year={2015}
}

@article{sukhbaatar2024btx,
  title={Branch-Train-MiX: Mixing Expert LLMs into a Mixture-of-Experts LLM},
  author={Sukhbaatar, Sainbayar and Goyal, Naman and Synnaeve, Gabriel and Lample, Guillaume},
  journal={arXiv preprint arXiv:2403.07816},
  year={2024}
}

@inproceedings{wan2024fusellm,
  title={Knowledge Fusion of Large Language Models},
  author={Wan, Fanqi and Huang, Xinting and Cai, Deng and Quan, Xiaojun and Bi, Wei and Shi, Shuming},
  booktitle=ICLR,
  year={2024}
}

@article{camkd2024,
  title={Confidence-Aware Multi-Teacher Knowledge Distillation},
  author={Zhang, Hailin and Chen, Defang and Wang, Can},
  journal={arXiv preprint arXiv:2201.00007},
  year={2022}
}

@article{ko2026reopold,
  title={REOPOLD: Reward-Based On-Policy Distillation with Mixture-Based Reward Clipping},
  author={Ko, Jongwoo and Park, Sungmin and Kim, Joohyung},
  journal={arXiv preprint arXiv:2603.11137},
  year={2026}
}

@inproceedings{lin2022truthfulqa,
  title={TruthfulQA: Measuring How Models Mimic Human Falsehoods},
  author={Lin, Stephanie and Hilton, Jacob and Evans, Owain},
  booktitle={Proceedings of the 60th Annual Meeting of the Association for Computational Linguistics (ACL)},
  pages={3214--3252},
  year={2022}
}

@inproceedings{lewis2020retrieval,
  title={Retrieval-Augmented Generation for Knowledge-Intensive {NLP} Tasks},
  author={Lewis, Patrick and Perez, Ethan and Piktus, Aleksandara and Petroni, Fabio and Karpukhin, Vladimir and Goyal, Naman and K{\"u}ttler, Heinrich and Lewis, Mike and Yih, Wen-tau and Rockt{\"a}schel, Tim and Riedel, Sebastian and Kiela, Douwe},
  booktitle={Advances in Neural Information Processing Systems (NeurIPS)},
  volume={33},
  pages={9459--9474},
  year={2020}
}

@inproceedings{guu2020retrieval,
  title={{REALM}: Retrieval-Augmented Language Model Pre-Training},
  author={Guu, Kelvin and Lee, Kenton and Tung, Zora and Pasupat, Panupong and Chang, Ming-Wei},
  booktitle={Proceedings of the 37th International Conference on Machine Learning (ICML)},
  pages={3929--3938},
  year={2020}
}

\end{document}